 \documentclass[pmlr,twocolumn]{jmlr} 

\usepackage{booktabs}
\usepackage[load-configurations=version-1]{siunitx} 

\theorembodyfont{\upshape}
\theoremheaderfont{\scshape}
\theorempostheader{:}
\theoremsep{\newline}

\jmlrvolume{ML4H Extended Abstract Arxiv Index}
\jmlryear{2020}
\jmlrsubmitted{2020}
\jmlrpublished{}
\jmlrworkshop{Machine Learning for Health (ML4H) 2020}

\title[A decision-making tool to fine-tune abnormal levels in the CBC tests]{
A decision-making 
tool to fine-tune abnormal levels \titlebreak in
the complete blood count tests
}

\author{%
\Name{Marta Avalos-Fernandez} \Email{marta.avalos-fernandez@u-bordeaux.fr}\\
\addr SISTM team, INRIA BSO, F-33405, Talence, France \\
\addr Univ. Bordeaux, INSERM, BPH U1219, F-33000 Bordeaux, France
\AND
\Name{Hélène Touchais} \Email{helene.touchais@u-bordeaux.fr}\\
\addr SISTM team, INRIA BSO, F-33405, Talence, France \\
\addr INP-ENSEEIHT, F-31000, Toulouse, France
\AND
\Name{Marcela Henríquez-Henríquez} \Email{marcela.henriquezh@integramedica.cl}\\
\addr ELSA Clinical Laboratory, IntegraMedica, part of BUPA, Santiago, Chile\\
\addr Cardio MR, Millennium Science Initiative Program, ANID, Santiago, Chile
}

\begin{document}

\maketitle

\begin{abstract}
The complete blood count (CBC) performed by automated hematology analyzers is one of the most ordered laboratory tests. It is a first-line tool for assessing a patient's general health status, or diagnosing and monitoring disease progression. When the analysis does not fit an expected setting,  technologists manually review a blood smear using a microscope. The International Consensus Group for Hematology Review published in 2005 a set of criteria for reviewing CBCs. Commonly, adjustments are locally needed to account for laboratory resources and  populations characteristics. 
Our objective is to provide a decision support tool to identify which CBC variables are associated with higher risks of abnormal smear and at which cutoff values. We propose a cost-sensitive Lasso-penalized additive logistic regression combined with  stability selection. Using simulated and real CBC data, we demonstrate that our tool correctly identify the true cutoff values, provided that there is enough available data in their neighbourhood. 

\end{abstract}
\begin{keywords}
Interpretability, Lasso, GAM, Imbalance, Population Health.
\end{keywords}

\section{Introduction}
\label{sec:intro}
The complete blood count (CBC) with leukocyte differential count performed by automated hematology analyzers is one of
the most ordered laboratory tests since used as first-line tool for assessing a patients general health status, diagnosing and monitoring disease progression and therapy. An automated hematology analyzer is a machine that rapidly counts and discriminates blood cells depending on its size, complexity or staining. When the analysis does not fit an expected setting, the machine triggers a warning flag. Then, technologists manually prepare and review a blood smear using a microscope. 
The manual technique is more efficient for the identification of cytological nuances, particularly in the presence of immature or abnormal cells. However, this is one of the most time-consuming hematology laboratory tests and requires a high degree of technical skill to minimize errors inherent in the subjectivity of the procedure. There is great interest in reducing the number of automated CBCs requiring manual blood smear reviews without sacrificing the quality of patient care \citep{Cle2017Blood}.

The International Consensus Group for Hematology Review published in  2005 a set of rules as criteria for reviewing CBCs \citep{Barnes05}. However, laboratories must adjust these general rules to their patients and machine characteristics. 
Commonly, adjustments are done accounting for the technical capabilities of hematology automatons, the population/sample distributions of the CBC data and permanently updated expert guidelines. 
The goal is to achieve the best trade-off between optimizing human and financial resources (i.e. minimizing the false positive rate) and optimizing the quality of patient care (i.e. minimizing the false negative rate, usually set in advance).

The manual smear is also considered as an internal quality control procedure for the
evaluation of the automated hematology machine parameters. Regularly, all
(flagged and unflagged) samples produced during a limited period of time are analyzed to
determine the need for adjustments of the screening criteria for manual blood smear review.
These large amounts of data regularly produced could be used to train machine learning algorithms 
that could be integrated to the quality control procedure as a decision-making tool to fine-tune abnormal levels in the CBC tests.

The objective of the present work was to provide a machine learning tool for decision support in fine-tuning abnormal levels in the CBC tests at laboratory-level. Explicitly, we aim to identify which CBC variables  are associated with a higher risk of abnormal manual smear and at which cutoff values. This involves that the machine learning tool must provide interpretable  results. However, that should surely not be an argument for  sacrifying predictive performance.

Our proposal is based on cost-sensitive Lasso-penalized additive logistic regression. Additive functions are considered to belong to the space of piecewise constant functions. 
The natural sparsity encouraged by the Lasso penalty is combined with a stability selection procedure to enforce model stability \citep{Meinshausen10}. The rationale behind this choice is the following.
Manual smear results are the binary outcome (abnormal/normal). CBC tests data are the quantitative predictors. The logistic regression model is a popular choice in the biomedical field. 
Categorisation of continuous variables is common in clinical studies since 
results are easier to interpret. Thus, we apply additive models fitted by piecewise constant polynomials.  
Once the piecewise constant basis functions have been set, we have gone through in a linear space setting. 

Now, a key question is how to determine the optimal number and location of cutoff points. 
Standard variable selection procedures are applied when only one  or a couple of quantitative predictors are present \citep{Liquet2019CPMCGLM,Barrio2013Use}. 
Since the CBC data consist in some twenty biological measures, multiplied by the number of basis functions, the Lasso appears as a  valuable alternative  \citep{Tibshirani96}. 
To encourage stable results we used Stability selection: only variables frequently selected by the Lasso analysis over data subsamples are retained \citep{Meinshausen10}. 

Finally, normal manual smear blood tests are much more usual than abnormal tests (less than $10\%$ in the present study). We are faced with an imbalanced learning problem. 
The two main families of methods in imbalanced binary classification are sampling  and cost-sensitive methods \citep{He09}. 
A simple cost-sensitive strategy consists in weighting individuals' contributions to the likelihood function 
to account for the degree of disequilibrium or importance \citep{Dmochowski10}.


\section{Method}
\label{Method}
Let $\{(X_{i1},\ldots,X_{ip},Y_{i})\}_{i=1}^n$ be an i.i.d. $n$-sample and $\{(x_{i1},\ldots, x_{ip}, y_{i})\}_{i=1}^n$  a realization of the sample. $Y$ is binary (0 for normal, 1 for abnormal smears). 
Let us note 
$\mathbf{x}_j = (x_{1j},\ldots, x_{nj})^\top$.
Consider the additive logistic model: $
      \mathrm{ logit}(P(Y_{i}=1| {x}_{i1},\ldots,{x}_{ip}))=
      \beta_0+$
      $\sum_{j=1}^p f_j({x}_{ij})
      $, where $f_j: \mathbb{R}\rightarrow\mathbb{R}$ are unknown centred  
functions and $\beta_0$ the intercept. Let  $\{\chi^k_j(\mathbf{x}_j)\}_{k=1}^{p_j}$ be a fixed basis of functions and denote $\boldsymbol\chi^k_j=\chi^k_j(\mathbf{x}_j)$. Any $f_j$ can be expanded in terms of these basis and the unknown parameters $\{\beta^k_j\}_{k=1}^{p_j}$: 
$f_j(\mathbf{x}_j) = \sum_{k=1}^{p_j} \beta^k_j \boldsymbol\chi^k_j=\boldsymbol\chi_j\boldsymbol \beta_j$.
Consider now the particular case of piecewise constants $\chi^k_{ij}=1$ if $x_{ij}>q_{j}^k$ and 0 otherwise, with $q_{j}^k$ the $k$-th value of a collection of $K_j$ fixed values. Considering the negative log-likelihood loss function, we can write the weighted penalized optimization problem as:
 \begin{equation}
     \begin{array}{l}
      \displaystyle{
      \underset{\boldsymbol \beta}{\mathrm{max}} \left\lbrace
      \sum_{i=1}^{n}\omega_i\ln
\frac{e^{y_i{\sum_{j=1}^p \boldsymbol\chi_{ij}\boldsymbol \beta_j}}}{1+e^{{\sum_{j=1}^p \boldsymbol\chi_{ij}\boldsymbol \beta_j}}}
      -\lambda\| \boldsymbol\beta\|_{1}
\right\rbrace
}
      \enspace,
     \end{array}
\label{likelihood}
     \end{equation}
where 
the weights $\omega_i = \omega>1$ if $y_i = 1$ and $1$ if $y_i = 0$  are used to account for the degree of imbalance in the minority class.
$\lambda>0$ controls regularization. 
The intercept is omitted. 
The selection probability is computed and used as a continuous measure of the stability associated to the Lasso estimates 
$\widehat{\beta^k_j}$. 

\cite{Amato16} conducted a comprehensive review of parsimonious additive models and reformulated the estimation problem in terms of group Lasso. 
Sparse group Lasso and overlapping group Lasso are alternative reformulations \citep{Chouldechova15,Lou2016Sparse}. 
Our proposal is close to these approaches. Nevertheless, by choosing piecewise constant functions, which conveniently categorises CBC data, interpretation, estimation, and computational issues  are dramatically simplified. 
The fused Lasso has also been adapted to multiple change-point detection by piecewise constant functions \citep{Petersen2019Data}. However, this approach is not well suited for large data.

\subsection{Evaluation}
\label{evaluation}
CBCs and blood smear reviews for 9~594 patients performed at the clinical laboratory of the Pontificia Universidad Catolica de Chile in 2016 were available.
CBC data consisted in 
hemoglobin (Hg in $g/dL$), 
hematocrit (Ht in $\%$), 
mean corpuscular hemoglobin concentration (MCHC in $g/dL$), 
mean corpuscular volume (MCV in $fL$), 
erythrocytes (Er in $10^6/\mu L$),
platelets (P in $10^3/\mu L$), 
red blood cell distribution width (RDW-CV in $\%$),
leukocytes (Le in $10^3/\mu L$), 
immature granulocytes (IG in $\%$) and 
the leukocyte differential count which includes 
neutrophils (N), basophils (B), eosinophils (Eo), monocytes (M) and lymphocytes (Ly) (in $10^3/\mu L$ and $\%$). 
Alarms of suspected alterations of blood cells (binary), sex (binary) and age (in years) were also reported. 
The response, normal/abnormal smear, is imbalanced: only $7\%$ of the smears were abnormal.

Real  data served as the core of a simulation study to evaluate if our procedure was able to detect relevant predictors and relevant cut-off values. The features were defined by the original data. $Y$ was  generated 
assuming the logistic model. 
We considered 4 scenarios: A and B, with only 2 (low-correlated) relevant predictors and C and D with 5 relevant predictors which shared moderate to high correlation between them and low correlation with all the other variables. 
Only one $\beta^k_j$ per variable was different to $0$. In scenarios A and D, they were placed in the extreme percentiles, where observations are scarce. Inversely, in scenarios B and C, they were placed in frequently observed values.  
$\beta_0$ was calibrated to achieve $7\%$ of events.
$Y$ was generated 100 times
 and selection probabilities computed.

\section{Results}
In all the 4 scenarios, irrelevant variables were infrequently selected  (Figure \ref{fig:f4}). 
In scenarios A and D, in which the true features have to be learn from scarce examples, their variability is slightly higher and the target value for the true cut-off values are less often achieved. 
Indeed, in scenario A, the 2 true values are identified as non zero in $52\%$ and $68\%$ of the situations, respectively. 
In scenario D, the 5 true values are identified as non zero in few situations. The procedure doesn't present good identification performance when relevant predictors are correlated to other relevant predictors and the true cut-off values are placed in the extreme percentiles. 
Inversely, in scenarios B and C, the capability  to detect the real cut-off values is excellent.  Values in the neighbourhood of the real cut-off values are most frequently selected than other irrelevant values. The impact of this error on the selection of cut-off values is minor.

\section{Conclusion}
\label{Discussion}
No multivariate analysis studies, even based on classical statistical methods, have been used to adjust the consensus cut-off values.
In general, studies in the literature propose new cut-off values and compare them to the consensus ones using some criteria such as recall. However, they do not describe the process of deducing the new (more efficient) cut-off values. Likely, this process is based on acquired expertise which is not reproducible. 
Our procedure represents a good trade off between traditional interpretable models \citep{Vellido2019Importance}  
and powerful machine learning methods for  clinical laboratories. It is able to detect true cut-off values with high probability, provided that enough data are available in their neighbourhood, while comparable in predictive performance to deep learning (results showed in \cite{Avalos2020Optimising}).  

Our tool has a straightforward practical application from an open source code available on request from the corresponding author (shortly on \url{https://github.com/mavalosf}).
\begin{figure}[tb]
\floatconts
  {fig:f4}
 {\caption{ Selection probability 
 for the 4 simulated scenarios. Values are mean over 100 simulation. 
 Black and gray crosses indicate estimated values of relevant and irrelevant predictors, respectively. Circles indicate target values. }}
{ \includegraphics[angle=90,width=0.95\linewidth]{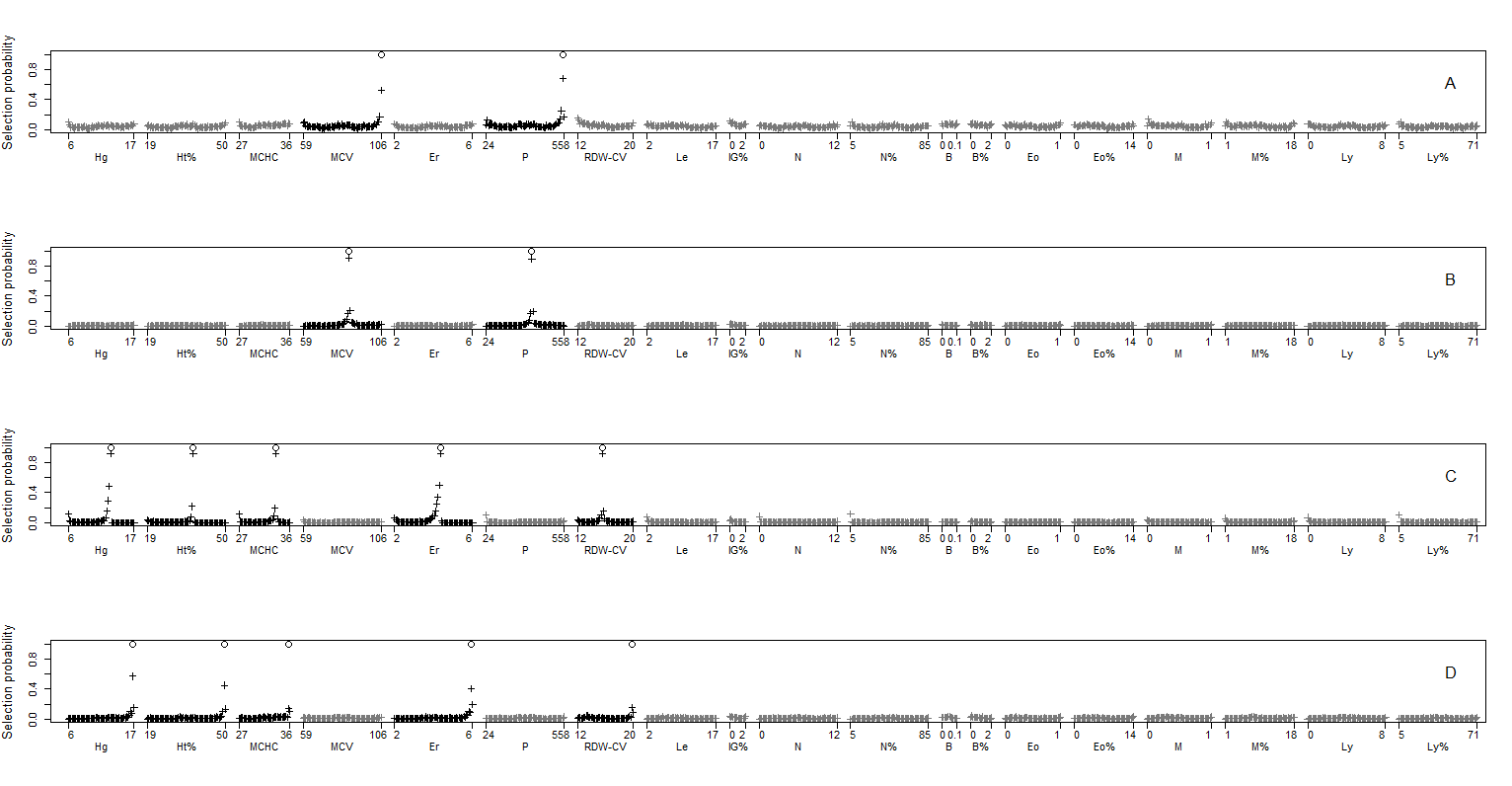}}
\end{figure}

\acks{
The ideas of this work were born from informal discussions between Marta Avalos and Marcela Henriquez during their research visit in Canberra, Australia (CSIRO's Data61 and ANU John Curtin School of Medical Research, respectively) in 2017. 
Ideas took shape during the participation of Marta Avalos to the "Big data: the information revolution in biomedical research" symposium organised by Marcela Henriquez and her colleagues from the Pontificia Universidad Catolica de Chile in December 2018.

Marta Avalos kindly thanks Cheng Soon Ong (CSIRO's Data61, Canberra) and Aditya Krishna Menon (formerly, CSIRO's Data61, Canberra; currently, Google, New York) for sharing their knowledge on imbalanced class learning.
Marcela Henriquez is indebted to Teresa Quiroga from the Department of Clinical Laboratories, School of Medicine, Pontificia Universidad Catolica de Chile, Santiago, Chile for contributions regarding the CBC data and for advice.

The calculations were carried out on the CURTA cluster of Mesocenter for Intensive Calculation in the Aquitaine French region. 
}

\bibliography{refs}

\appendix

\section{Appendix}\label{apd:first}
Supplementary figures are presented in the appendix.
\begin{figure}[tb]
\floatconts
  {fig:scheme}
 {\caption{ Heatmap of absolute value correlations of continuous predictors in the real CBC data. }}
{ \includegraphics[width=0.95\linewidth]{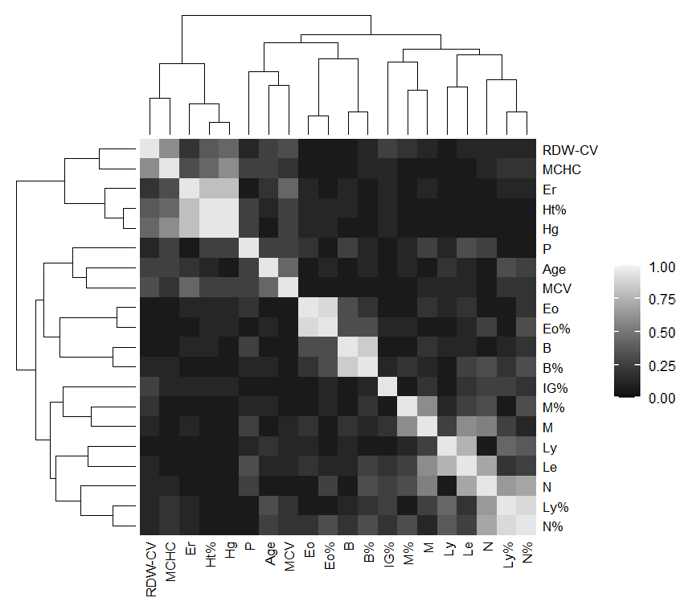}}
\end{figure}

\begin{figure}[tb]
\floatconts
  {fig:scheme}
 {\caption{ Scheme of the clinical laboratory procedures. }}
{ \includegraphics[angle=90,width=1.1\linewidth]{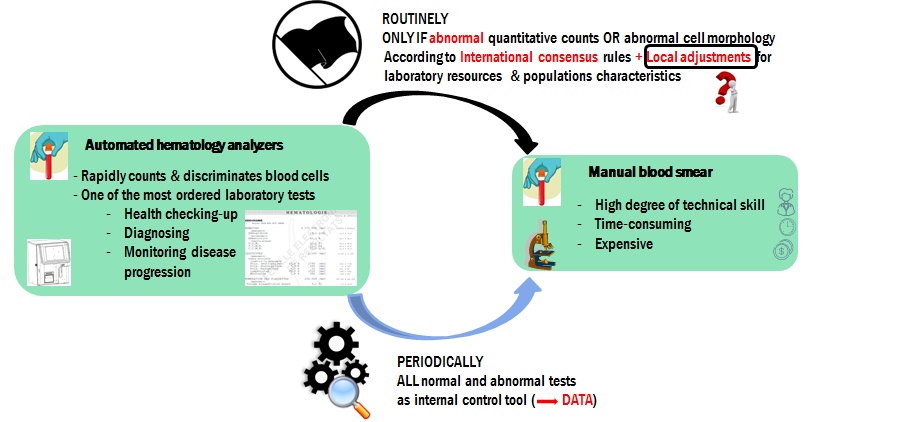}}
\end{figure}

\end{document}